\pdfoutput=1

\documentclass[11pt]{article}

\usepackage{acl}

\usepackage{times}
\usepackage{latexsym}
\usepackage{makecell}
\usepackage[T1]{fontenc}

\usepackage[utf8]{inputenc}

\usepackage{microtype}
\usepackage{breakurl}
\usepackage{times}
\usepackage{latexsym}

\usepackage{microtype}
\usepackage{hyperref}

\usepackage{graphicx}
\usepackage{xspace}
\usepackage{paralist}
\usepackage{booktabs}
\usepackage{multirow}
\usepackage{amsmath}
\usepackage{amsfonts}
\usepackage{amssymb}
\usepackage{pifont}
\usepackage{mathrsfs}
\usepackage{algorithm,algpseudocode}
\usepackage{mdwlist}
\usepackage{enumitem}
\usepackage{color}
\usepackage{dsfont}
\usepackage{amsthm}
\usepackage{bm}
\usepackage{float}
\usepackage{caption}
\usepackage{subcaption}
\usepackage{hyperref}
\usepackage{tcolorbox}
\usepackage{tikz}
\newcommand*{\circled}[1]{\lower.7ex\hbox{\tikz\draw (0pt, 0pt)%
    circle (.5em) node {\makebox[1em][c]{\small #1}};}}

\newcommand{\notes}[1]{}

\theoremstyle{definition}

\theoremstyle{plain}

\newcommand{\ith}[1]{\ensuremath{i^{{th}}}}

\newcount\permx
\newcount\permy
\def\permdot#1#2{
\permx=#1 \advance\permx by-1
\permy=#2 \advance\permy by-1
\psframe[fillcolor=black, fillstyle=solid]
(\permx,\permy)(#1, #2)
}

\newcommand{\boxnum}[1]{{\setlength{\fboxsep}{1pt}\raisebox{1pt}{\hspace{1pt}\fbox{\tiny #1}\hspace{1pt}}}}
\newcommand{\ind}[1]{\ensuremath{_{\kern-0.5pt\boxnum{#1}}}}

\newcommand{\smallnt}[1]{\ensuremath{_{\mbox{\tiny PP}}}\xspace}

\newcommand{\pseudocode}{Algorithm}
\floatname{algorithm}{\pseudocode}

\iffalse

\else

\fi

\newcommand{\RNum}[1]{\uppercase\expandafter{\romannumeral #1\relax}}

\begin{document}

\renewcommand{\thefootnote}{\alph{footnote}}

\title{SpeechAgents: Human-Communication Simulation with\\ Multi-Modal Multi-Agent Systems}

\author{
    \textbf{Dong Zhang\textsuperscript{\rm 1}},
    ~\textbf{Zhaowei Li\textsuperscript{\rm 2}},
    ~\textbf{Pengyu Wang\textsuperscript{\rm 1}},
    ~\textbf{Xin Zhang\textsuperscript{\rm 1}},
    \\
    ~\textbf{Yaqian Zhou\textsuperscript{\rm 1}}\thanks{~~Corresponding author},
    ~\textbf{Xipeng Qiu\textsuperscript{\rm 1}}\footnotemark[\value{footnote}] \\
    \textsuperscript{\rm 1}School of Computer Science, Fudan University\\
    \textsuperscript{\rm 2}School of Information Science and Technology, Fudan University\\
    {\tt 	dongzhang22@m.fudan.edu.cn} \\
    {\tt 	\{zhouyaqian,xpqiu\}@fudan.edu.cn} 
    \\
    \url{https://0nutation.github.io/SpeechAgents.github.io/}
}
\maketitle
\maketitle

\renewcommand{\thefootnote}{\fnsymbol{footnote}}

\setcounter{footnote}{0}
\renewcommand{\thefootnote}{\arabic{footnote}}

\begin{abstract}
Human communication is a complex and diverse process that not only involves multiple factors such as language, commonsense, and cultural backgrounds but also requires the participation of multimodal information, such as speech. Large Language Model~(LLM)-based multi-agent systems have demonstrated promising performance in simulating human society. Can we leverage LLM-based multi-agent systems to simulate human communication?
However, current LLM-based multi-agent systems mainly rely on text as the primary medium.
In this paper, we propose SpeechAgents, a multi-modal LLM based multi-agent system designed for simulating human communication. 
SpeechAgents utilizes multi-modal LLM as the control center for individual agent and employes multi-modal signals as the medium for exchanged messages among agents. Additionally, we propose Multi-Agent Tuning to enhance the multi-agent capabilities of LLM without compromising general abilities. To strengthen and evaluate the effectiveness of human communication simulation, we build the Human-Communication Simulation Benchmark. Experimental results demonstrate that SpeechAgents can simulate human communication dialogues with consistent content, authentic rhythm, and rich emotions and demonstrate excellent scalability even with up to 25 agents, which can apply to tasks such as drama creation and audio novels generation. Code and models will be open-sourced at \url{https://github.com/0nutation/SpeechAgents}.
\end{abstract}

\section{Introduction}
Human communication is a complex and diverse process involving various factors such as language, emotions, non-verbal expressions, and cultural backgrounds~\cite{devito2018human}. It also encompasses multiple modalities, such as speech~\cite{holler2019multimodal}. Utilizing artificial intelligence for simulating human communication can enhance our understanding of the essence of language and interaction, enabling the exploration of cognitive processes and social mechanisms in human society~\cite{troitzsch2012simulating}.
Current simulation systems for multi-modal human communication often focus on the modality extension but failed to generate high-quality dialogue content without relying on additional textual references~\cite{nguyen2022generative, mitsui2023humanlike}.
Leveraging the powerful understanding and generation capabilities of large language models~(LLM)~\cite{openai2023gpt4,touvron2023llama}, LLM-based multi-agent systems~\cite{li2023camel, talebirad2023multiagent, chen2023agentverse} has demonstrated promising performance in simulating human society~\cite{park2023generative}, historical events~\cite{hua2023war}, and debating~\cite{chan2023chateval}. Can we use LLM-based multi-agent systems to simulate multi-modal human communication?

\begin{figure}[t]
\centering
\includegraphics[width=1\columnwidth]{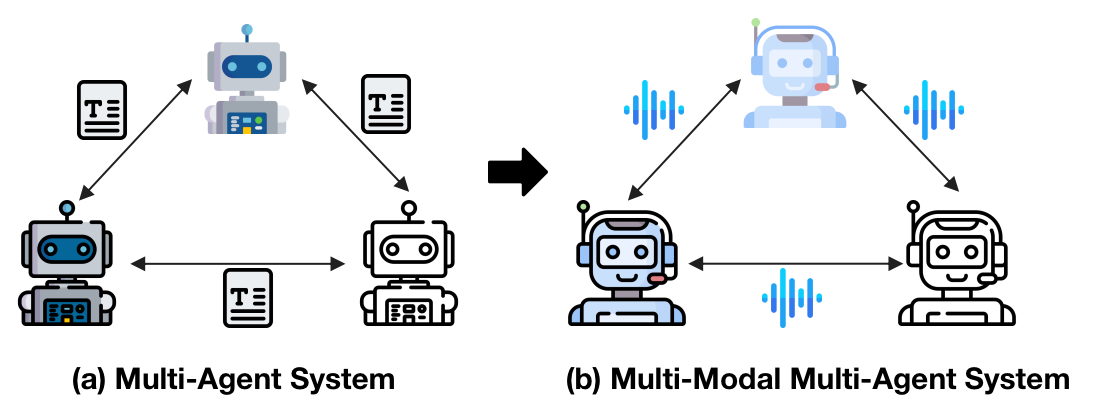} 
\caption{(a)~LLM-based Multi-Agent System is built on text-based LLM and rely on text as the medium for information exchange. (b)~Multi-modal LLM-based Multi-Agent System is built on multi-modal LLM and rely on multi-modal signals as the medium for information exchange}
\label{fig:mas}
\end{figure}

However, current LLM-based multi-agent systems employ text-based LLM as the central control and utilize text as the medium for information exchange among agents~\cite{qian2023communicative, hong2023metagpt, talebirad2023multiagent}, as shown in Figure~\ref{fig:mas}~(a). Consequently, they lack the capability to perceive and generate multi-modal signals.
Current multi-modal agents primarily utilize text-based LLM as the central control hub, interacting with other modalities through tool use of modality-specific experts~\cite{shen2023hugginggpt,yang2023mmreact,wu2023visual,huang2023audiogpt}. In such system, multi-modal capabilities are not inherently ingrained in agents, unlike text, posing challenges for seamless information integration and knowledge transfer across modalities.
Meanwhile, current exploration of multi-modal agents focus on individual agents~\cite{li2023multimodal}, lacking exploration into the construction of a multi-modal LLM-based multi-agent system .
 

We propose SpeechAgents, a multi-modal LLM based multi-agent system designed to simulate human communication. Concretely, we adopt SpeechGPT~\cite{zhang-etal-2023-speechgpt}, a multi-modal LLM that supports multi-modal input and output, as the control centor for individual agent. Different agents communicate with each other through speech signals.
To enhance and evaluate the multi-modal human communication simulation capabilities, we introduce the Human-Communication Simulation Benchmark. We propose multi-agent tuning to improve the multi-agent capabilities of the LLM without compromising general abilities. Experimental results demonstrate that SpeechAgents can generate human-like communication dialogues with accurate content, authentic rhythm, and rich emotions and demonstrate
excellent scalability even with up to 25 agents, which can apply to tasks such as drama creation and audio novels generation.

Our contributions include the following:
 \begin{itemize}[itemsep=1pt, leftmargin=10pt, parsep=0pt, topsep=1pt]
    \item 
    We build a multi-modal LLM based multi-agent system for human communication simulation and demonstrate the effectiveness of multi-modal signals as the medium of information exchange between agents.

    \item 
    We propose Multi-Agent Tuning to enhance the multi-agent capabilities of LLM without compromising general abilities.

    \item 
    We introduce the Human-Communication Simulation Benchmark.

\end{itemize}

\begin{figure*}[ht] 
    \setlength{\abovecaptionskip}{-0.cm}
    \centering 
    \includegraphics[width=1\textwidth]{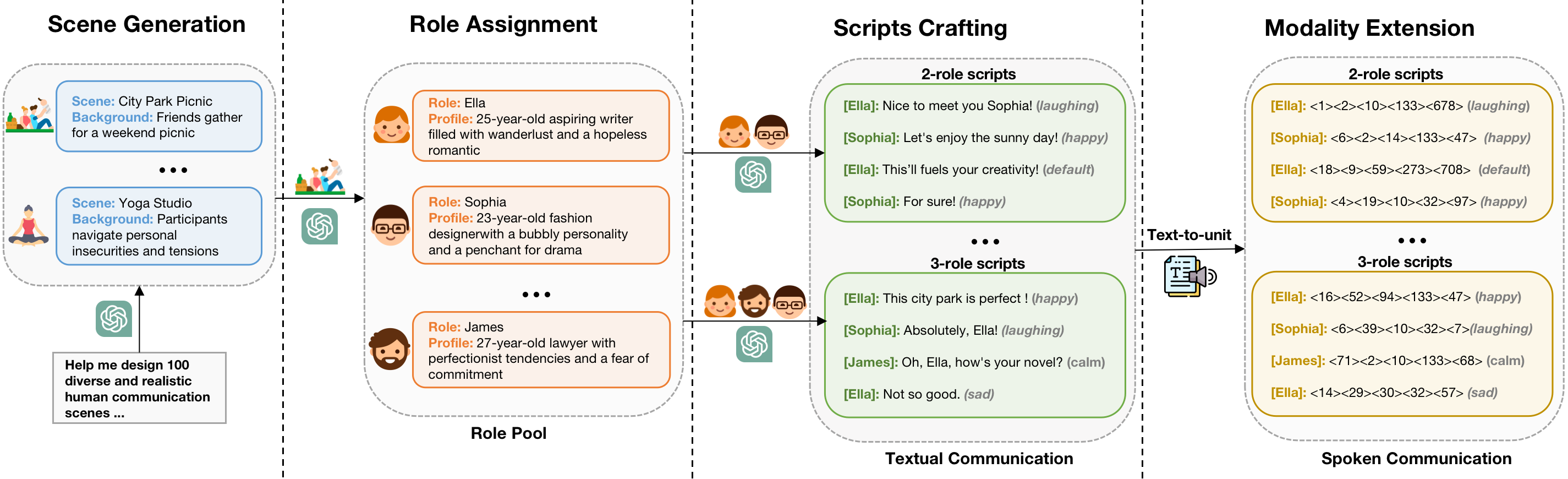} 
    \captionsetup{skip=10pt}
    \caption{An overview of Hmuan-Communication Simulation Benchmark construction process. We initiate the process by creating diverse scenes that simulate human communication. Subsequently, a role pool containing various roles is generated for each scene. Roles are then selected from the pool, and communication scripts are generated, depending on the specific scene and roles involved. Ultimately, multi-modal human communication scripts are crafted through text-to-speech conversion.}

    \label{fig:bench} 
\end{figure*}

\section{Related Work}
\noindent\textbf{Human-Communication Simulation}~
Several studies have explored the generation of human-like dialogues. For instance, dGSLM~\cite{nguyen2022generative} autonomously generates two-channel spoken dialogues, demonstrating realistic interactions between agents, including vocal interactions, laughter, and turn-taking. Similarly, CHATS~\cite{mitsui2023humanlike} transforms written dialogues into spoken form, ensuring coherence with the input text while introducing backchannels, laughter, and smooth turn-taking.
However, these systems mentioned above fall short in producing high-quality content without additional textual reference. In SpeechAgents, we leverage the powerful text comprehension and generation capabilities of LLM and build a multi-modal LLM SpeechGPT~\cite{zhang-etal-2023-speechgpt} based multi-agent system, which can generate multimodal signals while producing high-quality content. This advantage enables its application to tasks like drama creation and audio novels generation.

\noindent\textbf{Multi-Agent System}~
A Multi-Agent System (MAS) consists of multiple intelligent agents that collaboratively formulate decisions and execute corresponding actions in a distributed and parallel manner, significantly enhancing work efficiency and effectiveness~\cite{stone2000multiagent}. 
Currently, numerous LLM-based MASs are employed to accomplish complex tasks or simulate real-world scenarios. One noteworthy example is CAMEL~\cite{li2023camel}, a role-playing communicative agent framework that incorporates scenarios where two agents engage in interactive role-playing, showcasing the system's potential in addressing complex real-world situations. Another notable MAS involves a generative agent framework within a West World simulation~\cite{park2023generative}, introducing agents capable of mimicking human behavior in an interactive sandbox environment. 
However, existing MASs predominantly rely on text as the information carrier~\cite{talebirad2023multiagent, chen2023agentverse}, lacking effective processing and utilization of speech or other modal signals. In SpeechAgents, we use multiple agents to communicate through multi-modal signals.

\noindent\textbf{Multi-Modal Agent}~
Current multi-modal agents typically use text-based LLM as the central control, enhancing language-only models like ChatGPT~\cite{openai2023gpt4} with various multi-modal tools. Leveraging the robust knowledge base and reasoning capabilities of LLM, these agents can successfully tackle a variety of complex multi-modal tasks. For example, Visual ChatGPT~\cite{wu2023visual} facilitates dialogue-based image editing by integrating various image generation tools. MM-ReAct~\cite{yang2023mmreact} demonstrates that by collaborating with advanced vision experts, ChatGPT can execute complex multi-modal actions and reasoning. AudioGPT~\cite{huang2023audiogpt} extends ChatGPT's capabilities by incorporating audio foundation models to handle complex audio tasks.
However, the exploration of multi-modal agents predominantly focus on single-agent scenarios, lacking investigations into the construction of multi-agent systems. In SpeechAgents, we develop a multi-agent system based on a multi-modal LLM, SpeechGPT~\cite{zhang-etal-2023-speechgpt}, to simulate Human-Communication interactions, demonstrating the potential of a multi-modal LLM-based approach in achieving realistic human-like communication simulations.

\section{Hmuan-Communication Simulation Benchmark}

Human communication is an exceedingly diverse phenomenon, characterized by a wide range of scenarios, content, and participants. In order to enhance and evaluate the effectiveness of LLM-based agents in simulating human communication, we develop Human-Communication Simulation Benchmark, as illustrated in Figure~\ref{fig:bench}. We employ ChatGPT~(GPT-3.5-turbo) to generate human communication data hierarchically at three levels: \textit{scene}, \textit{role}, and \textit{scripts}.  Finally, we extend the modality of the data from text to speech through modality extension.

\noindent\textbf{Scene Generation}~
Scenes serve as specific locations where communication activities take place. We employ the zero-shot approach by prompting ChatGPT to generate various communication scenes, each with unique story backgrounds. This involves providing detailed descriptions of the time and location, as well as overall atmosphere to ensure the model can produce imaginative and diverse stories across various contexts. Detailed prompts are shown in Appendix~\ref{sec:app:prompts_scene_gen}. We generated 300 scenes for the training set and 50 scenes for the test set. Examples of generated scenes are listed in Appendix~\ref{sec:app:example_scene_gen}.

\noindent\textbf{Role Assignment}~
Character portrayal plays a crucial role in simulating human communication. For each specific scene, we prompt ChatGPT to create a diverse \textit{role pool} comprising 30 named characters, each accompanied by a brief description detailing their age, background, personality, and current state. Detailed prompts are shown in Appendix~\ref{sec:app:prompts_role_gen}. When generating dialogue scripts for particular scenarios, we can randomly select character candidates from this pool, adding variation and depth to the conversational scenes. Examples of generated roles are listed in Appendix~\ref{sec:app:example_role_gen}.

\noindent\textbf{Scripts Crafting}~
After determining the communication scene and background, we begin by randomly sampling a specific number of roles from the role pool, which will be used to generate dialogue scripts. We set the role number to 2, 4, 6, 8, and 10. Subsequently, we instruct ChatGPT to generate communication scripts that adhere to these specified conditions. Detailed prompts are shown in Appendix~\ref{sec:app:prompts_scripts_gen}. These scripts take the form of multi-party, multi-turn dialogues, ensuring that the dialogue content aligns with the scene description and that each character's speech corresponds to their personal profile. We require the dialogues to be logically consistent, contextually relevant, and rich in content. To enhance the simulation's realism, each character is expected to output the textual content and corresponding speaking style. Generated scripts examples are listed in Appendix~\ref{sec:app:example_scripts_gen}.

\noindent\textbf{Modality Extension}~
We aim to construct multi-modal human communication scripts, expanding communication scenarios from text to speech. As SpeechGPT utilizes discrete units as speech representation, we employ a pretrained text-to-unit generator\footnote{\url{https://huggingface.co/fnlp/text2unit}} to transform textual scripts into unit-form spoken scripts.

\section{SpeechAgents}
To simulate multi-modal human communication, we establish a Multi-modal Multi-Agent System. To enhance the multi-agent capabilities of the multi-modal LLM, we propose Multi-Agent Tuning.

\begin{figure*}[t] 
    \setlength{\abovecaptionskip}{-0.cm}
    \centering 
    \includegraphics[width=1\textwidth]{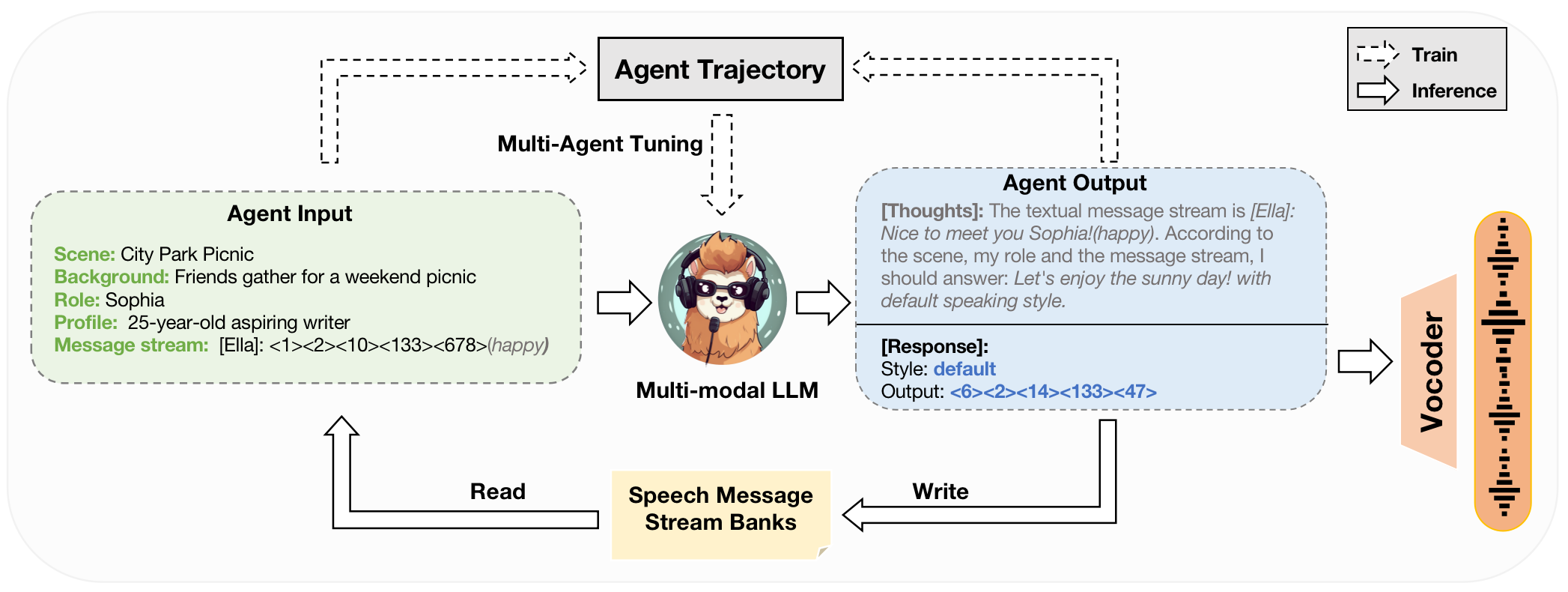} 
    \captionsetup{skip=10pt}
    \caption{Illustration of training and inference process of an individual agent in SpeechAgents. The solid arrows represent the data flow during the inference process. During one agent's turn, it receives inputs includes the scene, background, role, profile, and the message stream from the speech message stream banks. The agent's output consists of its inner thoughts, the generated speech response and corresponding style. The response with style is then written to the speech message stream bank. The dashed arrows represent the data flow during the training process. Agent trajectory instructions, parsed from scripts in the Human Communication Simulation Benchmark, are visually represented in the form of the concatenation of agent input and output in the diagram and utilized for multi-agent tuning of the multi-modal LLM.}
    \label{fig:trajectory} 
\end{figure*}


\subsection{Multi-modal Multi-Agent System}
\label{sec:041_mas}
The characteristics of multi-modal multi-agent system include: 1) Employing a multi-modal LLM as the central control unit for individual agents, and 2) Multimodal signals serve as the medium for communication among different agents, as shown in Figure~\ref{fig:mas}~(b).
We denote the set of agents in the system as $A$ and the set of messages as $M$.

\noindent\textbf{Multi-modal Agent}~
Each agent $i \in A$ is represented as $A_i = (L_i, S_i, R_i)$, where $L_i$ refers to the multi-modal LLM. The selection of the LLM can be decided by modality requirements. For instance, as we aim to extend human communication from text modality to speech, we choose the SpeechGPT series models as the central control for our agents. 
$S_i$ refers to the scene in which the agent is situated, including the corresponding background. $R_i$ denotes the role of the agent along with its associated profile. The scene and role guide the agent's actions and interactions.
In each round, the agent receives the message stream from other agents and generate appropriate an response consist with the scene and its role.

\noindent\textbf{Speech Message Stream}~
Agents communicate with each other through spoken interaction. Each agent's utterance serves as a message transmitted to all other agents. A \textit{speech message stream bank} is maintained to store the content of each participant's utterances in a spoken format.
Before each round, messages are retrieved from the message stream bank to inform the agent of what others have conveyed. After generating its response, it is then written into the message stream bank for reference in subsequent rounds.
Each message $m_{i,t} \in M$, sent from agent $A_i$ at turn $t$, can be represented as $m_{i,t} = (u_{i,t}, y_{i,t})$, where $u_{i,t}$ refers to the speech message and $y_{i,t}$ refers to the corresponding style.

\noindent\textbf{Think Before You Speak}~
When humans engage in communication, upon hearing others' words, they typically engage in internal thought processes before expressing their own opinions. Similarly, when each agent generates spoken output, we adhere to the principle of \textit{Think Before You Speak}. This approach is akin to the Chain-of-Thought (CoT) method, which has significantly enhanced the reasoning capabilities of LLM through step-by-step progress.
Specifically, we incorporate the guidance in the prompt: \textit{You should first think about the current condition and write your thoughts, and then output your response in this turn.} This instructs the agent to contemplate the present situation, formulate thoughts, and then articulate their response. Specifically, before an agent generates speech output, it should first create a textual message stream and then produce the corresponding text-based output, decomposing the complex task into several intermediate steps.

\noindent\textbf{Multi-Speaker Multi-Style Vocoder}~
To enhance the diversity and realism of simulated speech communication, we trained a multi-speaker multi-style vocoder following~\cite{nguyen2023expresso}. 
This vocoder takes speech discrete units, speaker, and style as inputs, producing speech with corresponding timbre and style. In each round, the output of each agent includes discrete units and the corresponding style, which are fed into the vocoder to generate expressive speech.
The vocoder architecture consists of a generator $\mathbf{G}$ and multiple discriminators $\mathbf{D}$. The generator uses look-up tables (LUT) to embed discrete representations and the embedding sequences are up-sampled by a series of blocks composed of transposed convolution and a residual block with dilated layers. 
The speaker embedding and style embedding is concatenated to each frame in the up-sampled sequence.
The discriminator features a Multi-Period Discriminator (MPD) and a Multi-Scale Discriminator (MSD), which have the same architecture as~\cite{nguyen2023expresso}.

\subsection{Multi-Agent Tuning}
\label{sec:042_mat}
To enhance the multi-agent capabilities of LLM, we introduce multi-agent tuning, similar to~\cite{zeng2023agenttuning}. Multi-agent tuning comprises two components: agent-trajectory instruction dataset derived from Human-Communication Simulation Benchmark dataset and a mix-tuning strategy. This strategy serves to augment the agent's multi-agent abilities while preserving its general capacity.

\noindent\textbf{Agent-Trajectory Parsing}~
Agent trajectory refers to the specific input and output corresponding to an individual agent, serving as training data for the agent's LLM. However, the training set in Human-Communication Simulation Benchmark consists of the input and output for the entire multi-agent system, not for individual agents. Consequently, it is necessary to parse the dataset into the format of agent trajectory.
In the Human-Communication Simulation Benchmark, each data pair can be represented as $(S, B, R, P, T_N, U_N)$, where:
\begin{itemize}
    \item $S$ and $B$ denote the scene and background,
    \item $R$ and $P$ represent the selected roles and corresponding profiles,
    \item $T_N$ refers to textual communication scripts containing $N$ round dialogues
    \item $U_N$ refers to spoken communication scripts containing $N$ round dialogues.
\end{itemize}

\noindent After parsing, each data point in the agent trajectory instruct-tuning dataset can be expressed as $(S, B, r, p, T_{i:j-1}, U_{i:j-1}, T_{j}, U_{j})$, where:
\begin{itemize}
    \item $r \in R$ and $p \in P$ denote the specific role and its profile for this turn, respectively.
    \item The textual message stream $T_{i:j-1}$ denotes the $i^{th}$ to $j-1^{th}$ round dialogue from $T_N$.
    \item The speech message stream $U_{i:j-1}$ denotes the $i^{th}$ to $j-1^{th}$ round dialogue from $U_N$.
    \item The textual output $T_{j}$ represents the $j^{th}$ turn dialogue of $T_N$.
    \item The speech output $U_{j}$ represents the $j^{th}$ turn dialogue of $U_N$.
\end{itemize}

\noindent After parsing all the data in the Human-Communication Simulation Benchmark, a total of 751,691 agent trajectories were obtained. 
Each agent trajectory will be fed into a template in Appendix~\ref{sec:app:template_agent_traj}, creating a sequence that will be utilized as the training data for multi-agent tuning.

\noindent\textbf{Mix-Tuning}~
We utilize the agent-trajectory instruction dataset to fine-tune the Language Model (LLM), enhancing the multi-agent ability of SpeechGPT. Simultaneously, we use Chain-of-Modality Instruction set of SpeechInstruct dataset\footnote{\url{https://huggingface.co/datasets/fnlp/SpeechInstruct}} to preserve the model's general ability.
The training objective for instruction tuning can be formated as:

\begin{equation*}
\begin{aligned}
L(\theta) = & -\cdot \mathbb{E}_{(x, y) \sim D_{\text{agent}}} [\log p(y | x)] \\
& - \alpha \cdot \mathbb{E}_{(x, y) \sim D_{\text{general}}} [\log p(y | x)]
\end{aligned}
\end{equation*}

where $D_{agent}$ denotes the agent-trajectory instruction dataset, $D_{general}$ denotes SpeechInstruct dataset and $\alpha$ represents the mixure ratio of $D_{agent}$ and $D_{general}$. We set $\alpha=1$.

\section{Experiments}
\subsection{Experimental Setups}
\label{sec:051_setup}
\textbf{Datasets}~
For multi-agent tuning, the agent-trajectory instruction dataset is parsed from Human-Communication Simulation Benchmark dataset. We also use Chain-of-Modality Instruction in SpeechInstruct dataset. For multi-speaker multi-style vocoder training, we use Expresso~\cite{nguyen2023expresso}, LJSpeech~\cite{ljspeech17} and VCTK dataset.

\noindent\textbf{Configuration}~
We train SpeechGPT from LLaMA2-7b-CHAT as the multi-modal LLM. We use the SpeechInstruct dataset and follow the stages of Cross-modal Instruction Fine-Tuning and Chain-of-Modality Instruction Fine-Tuning as described in~\cite{zhang-etal-2023-speechgpt}. We train for 77000 steps with batch size 1152 and maximum sequence length 1024 on 24 A100 GPUs. For multi-agent tuning, we train for 6000 steps with batch size 288 and maximum sequence length 4096 on 24 A100 GPUs. For decoding, we set the maximum
sequence length to 4096 and set the temperature to
0.8. We use Top-k sampling with k=60. We also
use Top-p sampling with p=0.8. 

\begin{table*}[]
\setlength{\belowcaptionskip}{-0.2cm}
    \renewcommand{\arraystretch}{1.2}
    \Large
    \centering
    \resizebox{1\textwidth}{!}{
    \begin{tabular}{l|cc|cc|cc|cc|cc|cc|c}
    \toprule
    \multirow{2}{*}{} & \multicolumn{12}{c}{Human-Communication Simulation} & General Ability \\
    & \multicolumn{2}{c}{2-Role} & \multicolumn{2}{c}{4-Role} & \multicolumn{2}{c}{6-Role} & \multicolumn{2}{c}{8-Role} & \multicolumn{2}{c}{10-Role} & 
    \multicolumn{2}{c}{Avg.} &\\
    Method &  C-Score & Q-Score &  C-Score  & Q-Score &  C-Score & Q-Score  &  C-Score  & Q-Score  &  C-Score & Q-Score  & C-Score & Q-Score & ChatGPT Score \\ 
    \midrule
    \multicolumn{12}{l}{\textit{Baselines}} \\ 
    Speech-ChatGPT  & 4.7 & 4.3 & 4.6  & 4.2  & 4.6  & 4.1  & 4.5  & 4.4  & 4.3  & 4.2 & \textbf{4.5} & \textbf{4.3} & -  \\
    LLaMA2-MAT &  4.4 & 3.8  & 4.3 & 3.8 & 4.1  & 3.6  & 4.2  & 3.8  & 4.2  & 3.9 & 4.2 & 3.8 & - \\
    Speech-LLaMA2-MAT &  4.1 & 3.7  & 4.2 & 3.7 & 3.9  & 3.5  & 4.0  & 3.6  & 4.0  & 3.6 & 4.0 & 3.6 & - \\
    \midrule
    SpeechGPT & 1.0 & 1.0  & 1.0 & 1.0 & 1.0 & 1.0 & 1.0 & 1.0 & 1.0 & 1.0 & 1.0 & 1.0 & 3.6 \\
    SpeechAgents & 4.1  & 3.7  & 4.2 & 3.6 & 4.0 & 3.7  & 3.9 & 3.9  & 4.3 & 3.9 & \textbf{4.1} & \textbf{3.8} & \textbf{3.9}  \\
    \quad -Mix-Tuning   & 4.1  & 3.8  & 4.1 & 3.5 &  4.1 & 3.8  & 4.0  &  3.9 & 3.9  &  3.9  &4.0 & \textbf{3.8} &  1.0 \\
    \quad -\textit{Think Before You Speak}   &  1.0 & 1.0  & 1.0 & 1.0 & 1.0  & 1.0  & 1.0  &  1.0 & 1.0  &  1.0 &1.0 &1.0  & - \\
    \bottomrule
    \end{tabular}}
    \caption{Evaluation results of SpeechAgents under Human-Communication scenarios containing different role numbers and speech-to-speech dialogue performance which represents general ability.
    C-Score refers to Content Score. Q-Score refers to Quality Score. ChatGPT Score follows the same setting in~\cite{zhang-etal-2023-speechgpt}. 
    }
    \label{tab:main}
\end{table*}

\subsection{Baselines}
\label{sec:052_baseline}
\noindent\textbf{Speech-ChatGPT} is a multi-agent system built upon cascaded spoken conversational systems, consisting of off-the-shell ASR systems~\footnote{\url{https://openai.com/research/whisper}}, ChatGPT~(GPT-3.5-turbo) as well as off-the-shell TTS systems~\footnote{\url{https://platform.openai.com/docs/guides/text-to-speech}}. 

\noindent\textbf{LLaMA2-MAT} is a text-based multi-agent system. The single agent is built upon a large language model obtained by performing textual multi-agent tuning on LLaMA2-7B-chat using agent-trajectory instruction dataset in section~\ref{sec:042_mat}. Textual multi-agent tuning leverages textual message stream instead of speech message stream. Template for textual multi-agent tuning is shown in Appendix~\ref{sec:app:template_agent_traj_text}. All other settings remain consistent with those described in section~\ref{sec:042_mat}.

\noindent\textbf{Speech-LLaMA2-MAT} is a multi-agent system built upon cascaded spoken conversational system, consisting of off-the-shell ASR systems~\footnote{\url{https://openai.com/research/whisper}}, \textit{LLaMA2-MAT} as well as off-the-shell TTS systems~\footnote{\url{https://platform.openai.com/docs/guides/text-to-speech}}.

\subsection{Evaluation}
\label{sec:053_eval}
We evaluate two key capabilities of SpeechAgents: the ability to simulate human communication and general ability. For human communication simulation evaluation, we use test set in Human-Communication Simulation Benchmark and utilize ChatGPT~(GPT-4) as an evaluator, primarily evaluating the generated scripts from two perspectives: consistency with the scenario and characters, and the quality and logical coherence of the script content. As for general ability, we evaluate SpeechAgents based on its performance in speech-to-speech dialogue tasks, as described in~\cite{zhang-etal-2023-speechgpt}.

\noindent\textbf{Consistency Score} evaluates whether the scripts align with the provided scene and character descriptions and contextual elements such as time and atmosphere.
We leverage the off-the-shell ASR model in section~\ref{sec:052_baseline} to transform the speech scripts into its corresponding text, which is subsequently submitted for evaluation. We feed the prompt in Appendix~\ref{sec:app:prompt_consistency_score} to ChatGPT to score the model's outputs based on response quality, with scores ranging from 1 to 5. The higher score represents the better consistency.

\noindent\textbf{Quality Score} focuses on language quality, emotional expression, logical consistency, and overall reasonableness of each dialogue, evaluating whether the scripts are natural, fluent, and free from grammatical and lexical errors.
We leverage the pre-trained ASR model in section~\ref{sec:052_baseline} to transform the speech scripts into its corresponding text, which is subsequently submitted for evaluation. We feed the prompt in Appendix~\ref{sec:app:prompt_content_score} to ChatGPT to score the model's outputs based on response quality, with scores ranging from 1 to 5. The higher score represents the better quality.

\noindent\textbf{Spoken Dialogue Score}
To assess the general ability, we evaluate the performance of LLM in SpeechAgents on speech-to-speech instruction-following task proposed in \cite{zhang-etal-2023-speechgpt} and focus on the quality of dialogue content. The processing progress, test dataset and evaluation metrics are consistent with those described in \cite{zhang-etal-2023-speechgpt}.

\subsection{Main Results}
\label{sec:054_main}
Table~\ref{tab:main} presents the evaluation of human-communication simulation on different roles and speech-to-speech dialogue for general ability.
Comparing the performance of SpeechAgents and SpeechGPT in Human-Communication Simulation, it is observed that SpeechAgents exhibits a clear advantage across all role numbers. This highlights the effectiveness of multi-agent tuning in enhancing the model's multi-agent ability. Additionally, when contrasting their performance in spoken dialogue, SpeechAgents even outperforms SpeechGPT, indicating that general ability has not been compromised. Moreover, the multi-agent tuning employed for Human-Communication Simulation tasks also contributes to the improvement of general ability.

\noindent In comparison to LLaMA2-MAT, SpeechAgents achieved similar consistency and quality scores. This underscores the effectiveness and significant potential of using multi-modal signals as the medium for information exchange among agents.

\noindent Speech-ChatGPT performs best in Human-Communication Simulation, primarily due to ChatGPT's great language understanding and generation capabilities. SpeechAgents outperforms Speech-LLaMA2-MAT in both consistency and quality scores, indicating that when a cross-modal LLM possessing inherent speech capabilities serves as the central control for agent, it yields better results than agents relying on modality experts to get multi-modal capabilities. This highlights the advantage of intrinsic cross-modal knowledge transfer in achieving superior performance in a multi-modal setting.




\section{Analysis}
\begin{figure}[t]
    \centering
    \includegraphics[width=\linewidth]{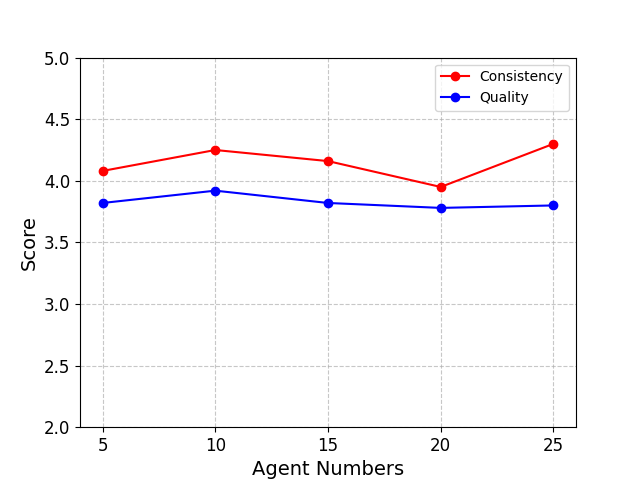}
    \caption{Consistency and Quality scores of SpeechAgents under Human-Communication scenarios containing different role numbers.}
    
    \label{fig:scale}
\end{figure}

\subsection{Ablation Study}
\label{sec:ablation}
\noindent\textbf{Effect of Mix-Tuning}~
Removing mix-tuning refers to finetuning multi-modal LLM solely on agent trajectory instructions without SpeechInstruct dataset. As shown in Table~\ref{tab:main}, the removal of Mix Training had no impact on the performance in Human-Communication Simulation. However, there was a significant loss in the performance in speech-to-speech dialogue. This indicates that Mix Training is highly effective in preserving general ability.

\noindent\textbf{Effect of \textit{Think Before You Speak}}~
Removing \textit{Think Before You Speak} means training and inference without \textit{Thoughts}. The template for removing \textit{Think Before You Speak} is shown in Appendix~\ref{sec:app:template_agent_traj_nothought}.
As indicated in Table~\ref{tab:main}, the removal of \textit{Think Before You Speak} essentially rendered the Human-Communication Simulation task unachievable. This highlights the critical significance of this design element.

\subsection{Scalability of Agent Numbers}
\label{sec:scale}

As depicted in Figure~\ref{fig:scale}, with the increase of agent numbers, the consistency score and quality score do not show a noticeable decrease, almost remaining unchanged. Even in a conversation scenario involving 25 agents, scripts with relatively high consistency and quality can be produced. This illustrates the SpeechAgents framework's strong scalability, demonstrating its ability to generalize to scenarios beyond the maximum agent number encountered in the training set.

\subsection{Case Study}
\label{sec:case}
The following lists an generated scripts. We also show the speech demo in our project page\footnote{\url{https://0nutation.github.io/SpeechAgents.github.io/}}. 
\begin{tcolorbox}[width=1\textwidth]
\textbf{Scene}: Summer Community BBQ\\
\textbf{Background}: Community members organize a summer barbecue, bringing people together for grilled food, music, and outdoor fun, fostering a sense of unity and camaraderie.\\
\textbf{Roles}:\\
  Michael Turner: 45-year-old grill master and friendly, taking charge of the barbecue station.\\
  Jessica Rodriguez: 30-year-old event planner and energetic, organizing the logistics of the community BBQ\\
  David Nguyen: 40-year-old local business owner and generous, sponsoring the event by providing refreshments.\\
  Olivia Simmons: 33-year-old parent and enthusiastic, organizing children's activities\\
  \\
\textbf{Generated scripts}:\\
    Michael Turner: Hey everyone, it's time for our summer community barbecue!(default)
  
    Jessica Rodriguez:Great job,Michael! I've made sure everything is set up perfectly. Now, let's start the barbecue!(excited)
    
    David Nguyen:I'm glad to see everyone enjoying the barbecue. It's a great opportunity for local businesses to connect with the community.(happy)
    
    Michael Turner: Thank you all for coming! I'm proud to see the community coming together to enjoy this barbecue.(default)
    
    Olivia Simmons: Now, let's have some fun with face painting and games!(happy)
    
    Olivia Simmons: kids, gather around! I've set up some face painting and games.(happy)
    
    Michael Turner: That sounds like a great idea, Olivia. I'll make sure to have some tasty snacks ready for everyone.(happy)
    
    Jessica Rodriguez: Great, thanks, Michael! I've already booked a caterer for the event. (happy)
    
    David Nguyen: That's fantastic, Jessica. I'm glad everyone is excited.(default)
    
    Olivia Simmons: Let's make sure we have plenty of space for the kids to play.(happy)
    
    Michael Turner: Absolutely, Olivia! I'll make sure the barbecue station is child-friendly.(happy)
    
    Jessica Rodriguez: Thank you, Michael. I've made sure to have a separate grilling area just for the kids.(default)
    
    David Nguyen: I'm grateful to have the opportunity to sponsor this event and support our community.(default)

\end{tcolorbox}

\section{Conclusion}
This work proposed SpeechAgents, a multi-modal LLM-based multi-agent system designed for simulating human communication. SpeechAgents utilize multi-modal LLM as the central control for agents and employing multi-modal signals as the medium for exchanged messages among agents. Experimental results demonstrate that SpeechAgents can simulate human communication dialogues with correct content, authentic rhythm, and rich emotions and demonstrate excellent scalability even with up to 25 agents, which can apply to tasks such as drama creation and audio novels generation.

\clearpage




\bibliography{custom}

\begin{thebibliography}{25}
\expandafter\ifx\csname natexlab\endcsname\relax\def\natexlab#1{#1}\fi

\bibitem[{Chan et~al.(2023)Chan, Chen, Su, Yu, Xue, Zhang, Fu, and Liu}]{chan2023chateval}
Chi-Min Chan, Weize Chen, Yusheng Su, Jianxuan Yu, Wei Xue, Shanghang Zhang, Jie Fu, and Zhiyuan Liu. 2023.
\newblock \href {http://arxiv.org/abs/2308.07201} {Chateval: Towards better llm-based evaluators through multi-agent debate}.

\bibitem[{Chen et~al.(2023)Chen, Su, Zuo, Yang, Yuan, Chan, Yu, Lu, Hung, Qian, Qin, Cong, Xie, Liu, Sun, and Zhou}]{chen2023agentverse}
Weize Chen, Yusheng Su, Jingwei Zuo, Cheng Yang, Chenfei Yuan, Chi-Min Chan, Heyang Yu, Yaxi Lu, Yi-Hsin Hung, Chen Qian, Yujia Qin, Xin Cong, Ruobing Xie, Zhiyuan Liu, Maosong Sun, and Jie Zhou. 2023.
\newblock \href {http://arxiv.org/abs/2308.10848} {Agentverse: Facilitating multi-agent collaboration and exploring emergent behaviors}.

\bibitem[{DeVito(2018)}]{devito2018human}
Joseph~A DeVito. 2018.
\newblock \emph{Human communication: The basic course}.
\newblock Pearson.

\bibitem[{Holler and Levinson(2019)}]{holler2019multimodal}
Judith Holler and Stephen~C Levinson. 2019.
\newblock Multimodal language processing in human communication.
\newblock \emph{Trends in Cognitive Sciences}, 23(8):639--652.

\bibitem[{Hong et~al.(2023)Hong, Zhuge, Chen, Zheng, Cheng, Zhang, Wang, Wang, Yau, Lin, Zhou, Ran, Xiao, Wu, and Schmidhuber}]{hong2023metagpt}
Sirui Hong, Mingchen Zhuge, Jonathan Chen, Xiawu Zheng, Yuheng Cheng, Ceyao Zhang, Jinlin Wang, Zili Wang, Steven Ka~Shing Yau, Zijuan Lin, Liyang Zhou, Chenyu Ran, Lingfeng Xiao, Chenglin Wu, and Jürgen Schmidhuber. 2023.
\newblock \href {http://arxiv.org/abs/2308.00352} {Metagpt: Meta programming for a multi-agent collaborative framework}.

\bibitem[{Hua et~al.(2023)Hua, Fan, Li, Mei, Ji, Ge, Hemphill, and Zhang}]{hua2023war}
Wenyue Hua, Lizhou Fan, Lingyao Li, Kai Mei, Jianchao Ji, Yingqiang Ge, Libby Hemphill, and Yongfeng Zhang. 2023.
\newblock \href {http://arxiv.org/abs/2311.17227} {War and peace (waragent): Large language model-based multi-agent simulation of world wars}.

\bibitem[{Huang et~al.(2023)Huang, Li, Yang, Shi, Chang, Ye, Wu, Hong, Huang, Liu, Ren, Zhao, and Watanabe}]{huang2023audiogpt}
Rongjie Huang, Mingze Li, Dongchao Yang, Jiatong Shi, Xuankai Chang, Zhenhui Ye, Yuning Wu, Zhiqing Hong, Jiawei Huang, Jinglin Liu, Yi~Ren, Zhou Zhao, and Shinji Watanabe. 2023.
\newblock \href {http://arxiv.org/abs/2304.12995} {Audiogpt: Understanding and generating speech, music, sound, and talking head}.

\bibitem[{Ito and Johnson(2017)}]{ljspeech17}
Keith Ito and Linda Johnson. 2017.
\newblock The lj speech dataset.
\newblock \url{https://keithito.com/LJ-Speech-Dataset/}.

\bibitem[{Li et~al.(2023{\natexlab{a}})Li, Gan, Yang, Yang, Li, Wang, and Gao}]{li2023multimodal}
Chunyuan Li, Zhe Gan, Zhengyuan Yang, Jianwei Yang, Linjie Li, Lijuan Wang, and Jianfeng Gao. 2023{\natexlab{a}}.
\newblock \href {http://arxiv.org/abs/2309.10020} {Multimodal foundation models: From specialists to general-purpose assistants}.

\bibitem[{Li et~al.(2023{\natexlab{b}})Li, Hammoud, Itani, Khizbullin, and Ghanem}]{li2023camel}
Guohao Li, Hasan Abed Al~Kader Hammoud, Hani Itani, Dmitrii Khizbullin, and Bernard Ghanem. 2023{\natexlab{b}}.
\newblock \href {http://arxiv.org/abs/2303.17760} {Camel: Communicative agents for "mind" exploration of large language model society}.

\bibitem[{Mitsui et~al.(2023)Mitsui, Hono, and Sawada}]{mitsui2023humanlike}
Kentaro Mitsui, Yukiya Hono, and Kei Sawada. 2023.
\newblock \href {http://arxiv.org/abs/2310.01088} {Towards human-like spoken dialogue generation between ai agents from written dialogue}.

\bibitem[{Nguyen et~al.(2023)Nguyen, Hsu, D'Avirro, Shi, Gat, Fazel-Zarani, Remez, Copet, Synnaeve, Hassid, Kreuk, Adi, and Dupoux}]{nguyen2023expresso}
Tu~Anh Nguyen, Wei-Ning Hsu, Antony D'Avirro, Bowen Shi, Itai Gat, Maryam Fazel-Zarani, Tal Remez, Jade Copet, Gabriel Synnaeve, Michael Hassid, Felix Kreuk, Yossi Adi, and Emmanuel Dupoux. 2023.
\newblock \href {http://arxiv.org/abs/2308.05725} {Expresso: A benchmark and analysis of discrete expressive speech resynthesis}.

\bibitem[{Nguyen et~al.(2022)Nguyen, Kharitonov, Copet, Adi, Hsu, Elkahky, Tomasello, Algayres, Sagot, Mohamed, and Dupoux}]{nguyen2022generative}
Tu~Anh Nguyen, Eugene Kharitonov, Jade Copet, Yossi Adi, Wei-Ning Hsu, Ali Elkahky, Paden Tomasello, Robin Algayres, Benoit Sagot, Abdelrahman Mohamed, and Emmanuel Dupoux. 2022.
\newblock \href {http://arxiv.org/abs/2203.16502} {Generative spoken dialogue language modeling}.

\bibitem[{OpenAI(2023)}]{openai2023gpt4}
OpenAI. 2023.
\newblock \href {http://arxiv.org/abs/2303.08774} {Gpt-4 technical report}.

\bibitem[{Park et~al.(2023)Park, O'Brien, Cai, Morris, Liang, and Bernstein}]{park2023generative}
Joon~Sung Park, Joseph~C. O'Brien, Carrie~J. Cai, Meredith~Ringel Morris, Percy Liang, and Michael~S. Bernstein. 2023.
\newblock \href {http://arxiv.org/abs/2304.03442} {Generative agents: Interactive simulacra of human behavior}.

\bibitem[{Qian et~al.(2023)Qian, Cong, Liu, Yang, Chen, Su, Dang, Li, Xu, Li, Liu, and Sun}]{qian2023communicative}
Chen Qian, Xin Cong, Wei Liu, Cheng Yang, Weize Chen, Yusheng Su, Yufan Dang, Jiahao Li, Juyuan Xu, Dahai Li, Zhiyuan Liu, and Maosong Sun. 2023.
\newblock \href {http://arxiv.org/abs/2307.07924} {Communicative agents for software development}.

\bibitem[{Shen et~al.(2023)Shen, Song, Tan, Li, Lu, and Zhuang}]{shen2023hugginggpt}
Yongliang Shen, Kaitao Song, Xu~Tan, Dongsheng Li, Weiming Lu, and Yueting Zhuang. 2023.
\newblock \href {http://arxiv.org/abs/2303.17580} {Hugginggpt: Solving ai tasks with chatgpt and its friends in huggingface}.

\bibitem[{Stone and Veloso(2000)}]{stone2000multiagent}
Peter Stone and Manuela Veloso. 2000.
\newblock Multiagent systems: A survey from a machine learning perspective.
\newblock \emph{Autonomous Robots}, 8:345--383.

\bibitem[{Talebirad and Nadiri(2023)}]{talebirad2023multiagent}
Yashar Talebirad and Amirhossein Nadiri. 2023.
\newblock \href {http://arxiv.org/abs/2306.03314} {Multi-agent collaboration: Harnessing the power of intelligent llm agents}.

\bibitem[{Touvron et~al.(2023)Touvron, Lavril, Izacard, Martinet, Lachaux, Lacroix, Rozi{\`e}re, Goyal, Hambro, Azhar et~al.}]{touvron2023llama}
Hugo Touvron, Thibaut Lavril, Gautier Izacard, Xavier Martinet, Marie-Anne Lachaux, Timoth{\'e}e Lacroix, Baptiste Rozi{\`e}re, Naman Goyal, Eric Hambro, Faisal Azhar, et~al. 2023.
\newblock Llama: Open and efficient foundation language models.
\newblock \emph{arXiv preprint arXiv:2302.13971}.

\bibitem[{Troitzsch(2012)}]{troitzsch2012simulating}
Klaus~G Troitzsch. 2012.
\newblock Simulating communication and interpretation as a means of interaction in human social systems.
\newblock \emph{Simulation}, 88(1):7--17.

\bibitem[{Wu et~al.(2023)Wu, Yin, Qi, Wang, Tang, and Duan}]{wu2023visual}
Chenfei Wu, Shengming Yin, Weizhen Qi, Xiaodong Wang, Zecheng Tang, and Nan Duan. 2023.
\newblock \href {http://arxiv.org/abs/2303.04671} {Visual chatgpt: Talking, drawing and editing with visual foundation models}.

\bibitem[{Yang et~al.(2023)Yang, Li, Wang, Lin, Azarnasab, Ahmed, Liu, Liu, Zeng, and Wang}]{yang2023mmreact}
Zhengyuan Yang, Linjie Li, Jianfeng Wang, Kevin Lin, Ehsan Azarnasab, Faisal Ahmed, Zicheng Liu, Ce~Liu, Michael Zeng, and Lijuan Wang. 2023.
\newblock \href {http://arxiv.org/abs/2303.11381} {Mm-react: Prompting chatgpt for multimodal reasoning and action}.

\bibitem[{Zeng et~al.(2023)Zeng, Liu, Lu, Wang, Liu, Dong, and Tang}]{zeng2023agenttuning}
Aohan Zeng, Mingdao Liu, Rui Lu, Bowen Wang, Xiao Liu, Yuxiao Dong, and Jie Tang. 2023.
\newblock \href {http://arxiv.org/abs/2310.12823} {Agenttuning: Enabling generalized agent abilities for llms}.

\bibitem[{Zhang et~al.(2023)Zhang, Li, Zhang, Zhan, Wang, Zhou, and Qiu}]{zhang-etal-2023-speechgpt}
Dong Zhang, Shimin Li, Xin Zhang, Jun Zhan, Pengyu Wang, Yaqian Zhou, and Xipeng Qiu. 2023.
\newblock \href {https://doi.org/10.18653/v1/2023.findings-emnlp.1055} {{S}peech{GPT}: Empowering large language models with intrinsic cross-modal conversational abilities}.
\newblock In \emph{Findings of the Association for Computational Linguistics: EMNLP 2023}, pages 15757--15773, Singapore. Association for Computational Linguistics.

\end{thebibliography}
\bibliographystyle{acl_natbib}

\clearpage
\appendix

\onecolumn\section{Prompt for Scene Generation}
\label{sec:app:prompts_scene_gen}
\begin{tcolorbox}[width=1\textwidth]
Help me design 100 diverse and realistic human communication scenes, each described in 20-40 words. Ensure that each scene is suitable for multiple participants, and each scene should not be complex. In the scene descriptions, please provide detailed depictions of the time and location while avoiding specifying the exact number of theatrical characters. The scenes should be different from each other and diverse.
\end{tcolorbox}

\section{Prompt for Role Assignment}
\label{sec:app:prompts_role_gen}
\begin{tcolorbox}[width=1\textwidth]
Please allocate 30 characters for this human communication scene, provide names and one-sentence profile for each character. The profile could include age, background, personality, and status. The profile must be a complete sentence, not words separated by commas. The descriptions should be within 10-30 words.  Organize the assignment in a python dict, with names as key and profiles as value. Ensure diverse character allocation and minimize duplications. You must allocate 30 roles. $\{scene\}$
\end{tcolorbox}

\section{Prompt for Scripts Crafting}
\label{sec:app:prompts_scripts_gen}
\begin{tcolorbox}[width=1\textwidth]
Please create a complete human communication scripts based on the scene description and character assignments. The scripts should consist of $\{n\_role\}$ characters.\\
\\
Here is the scene for the scripts: $\{scene\}$: $\{background\}$\\
\\
These are the characters and their profiles for the scripts in this scene: $\{roles\}:\{profiles\}$\\
\\
Please compose a scripts based on the scene and characters. The scripts should have 10-60 rounds of dialogue. The content of the scripts should fit the scene description, and each person's speech should match their profile. The scripts should be logically coherent, contextually consistent, and vivid in content. Each round should have only one speaker, and the speaker should be chosen from the given characters. The order of speakers is not fixed. Each person can speak up to 6 times. Each person's speech should fit the context and their role, and each person's dialogue should include the corresponding style. The style must be chosen from [default, happy, sad, angry, projected, calm, sleepy, bored, fearful, whisper, sympathetic, fast, desire, disgusted, confused, enunciated, laughing, sarcastic, narration, awe, animal, child, childdir, animaldir, nonverbal], and style choices can be diverse but must be consistent with the logical flow of the conversation. Your output must only include the dialogue content. The output needs to be saved in a Json file, with each line corresponding to the key: turn, role, style, content. 
\end{tcolorbox}

\clearpage
\section{Template for Agent Trajectory}
\label{sec:app:template_agent_traj}
\begin{tcolorbox}[width=1\textwidth]
Input:\\
Assuming you are an actor, currently joining a communication about $\{scene\}$. The background is $\{background\}$. You are $\{role\}$. Below is the description of your role: $\{profile\}$\\
\\
    You should first think about the current condition and write your thoughts, and then output your response in this turn. You should response in speech.\\
\\
    When responding, please output a response in the following format with two fields "Style" and "Output", your output must strictly follow this format:\\
    \\
    STYLE: (You should put the speaking style here)\\
    OUTPUT: (You should put what you want to speak use here)\\

    Here is the message streams:\\
    $\{speech\_message\_stream\}$\\
\\
    You should now give your response based on the above messages. Remember to give your response STRICTLY in the above response format. Do not add any additional field or line break to your response!\\
\\
Output:\\
$[$Thoughts$]$:\\
The textual message stream is $\{text\_message\_stream\}$. According to the scene, my role and message stream, I should answer $\{text\_output\}$ with $\{style\}$ speaking style.\\
\\

$[$Response$]$:\\
STYLE: $\{style\}$\\
OUTPUT: $\{speech\_output\}$\\
\end{tcolorbox}

\clearpage
\section{Template for Textual Agent Trajectory}
\label{sec:app:template_agent_traj_text}
\begin{tcolorbox}[width=1\textwidth]
Input:\\
Assuming you are an actor, currently joining a communication about $\{scene\}$. The background is $\{background\}$. You are $\{role\}$. Below is the description of your role: $\{profile\}$\\
\\

    When responding, please output a response in the following format with two fields "Style" and "Output", your output must strictly follow this format:\\
    \\
    STYLE: (You should put the speaking style here)\\
    OUTPUT: (You should put what you want to speak use here)\\

    Here is the message streams:\\
    $\{text\_message\_stream\}$\\
\\
    You should now give your response based on the above messages. Remember to give your response STRICTLY in the above response format. Do not add any additional field or line break to your response!\\
\\
Output:\\
$[$Response$]$:\\
STYLE: $\{style\}$\\
OUTPUT: $\{text\_output\}$\\
\end{tcolorbox}

\clearpage
\section{Template for Agent Trajectory Without Thoughts}
\label{sec:app:template_agent_traj_nothought}
\begin{tcolorbox}[width=1\textwidth]
Input:\\
Assuming you are an actor, currently joining a communication about $\{scene\}$. The background is $\{background\}$. You are $\{role\}$. Below is the description of your role: $\{profile\}$\\
\\
    You should response in speech.
    \\
    When responding, please output a response in the following format with two fields "Style" and "Output", your output must strictly follow this format:\\
    \\
    STYLE: (You should put the speaking style here)\\
    OUTPUT: (You should put what you want to speak use here)\\

    Here is the message streams:\\
    $\{speech\_message\_stream\}$\\
\\
    You should now give your response based on the above messages. Remember to give your response STRICTLY in the above response format. Do not add any additional field or line break to your response!\\
\\
Output:\\
$[$Response$]$:\\
STYLE: $\{style\}$\\
OUTPUT: $\{speech\_output\}$\\
\end{tcolorbox}

\clearpage
\section{Examples for Scene Generation}
\label{sec:app:example_scene_gen}
\begin{tcolorbox}[width=1\textwidth]

  City Park Picnic: Friends gather for a weekend picnic in the city park, discussing relationships, aspirations, and hidden conflicts.,
  \\
  \\
  Beach Bonfire: A group of friends shares stories around a bonfire on the beach, revealing secrets and challenging long-standing friendships.,
  \\
  \\
  Hospital Cafeteria: In the hospital cafeteria, healthcare workers cope with the stress of their jobs and confront ethical dilemmas.,
  \\
  \\
  Rural Farmhouse Kitchen: A family argues in the farmhouse kitchen over the future of the family farm, bringing generational conflicts to light.,
  \\
  \\
  Street Market in Marrakech: Vendors and tourists clash at a bustling street market in Marrakech, highlighting cultural misunderstandings and personal disputes.,
  \\
  \\
  Yoga Studio: Participants in a yoga class navigate personal insecurities and tensions, exploring the intersection of physical and emotional well-being.,
  \\
  \\
  Train Station Platform: Strangers waiting for a delayed train engage in conversations, exposing their diverse backgrounds and life stories.,
  \\
  \\
  Comedy Club Backstage: Comedians deal with backstage drama and rivalry before a stand-up show, shedding light on the cutthroat world of comedy.,
  \\
  \\
  Vineyard Harvest Celebration: At a vineyard celebration, workers and owners confront issues of fair wages and appreciation for their labor.\\

\end{tcolorbox}

\section{Examples for Role Assignment}
\label{sec:app:example_role_gen}
The following lists the 10 roles and profiles selected from the 30 roles generated for the "City Park Picnic" scene from Human Communication Benchmark.

\begin{tcolorbox}[width=1\textwidth]
  Victoria: 27-year-old journalist with a relentless pursuit for the truth and a turbulent love life.,
  \\
  \\
  Andrew: 29-year-old journalist with an adventurous spirit and a tendency to push boundaries.,
  \\
  \\
  Ella: 25-year-old aspiring writer filled with wanderlust and a hopeless romantic.,
  \\
  \\
  James: 27-year-old lawyer with perfectionist tendencies and a fear of commitment.,
  \\
  \\
  Sophia: 23-year-old fashion designer with a bubbly personality and a penchant for drama.,
  \\
  \\
  Oliver: 28-year-old architect with a mysterious past and a guarded demeanor.,
  \\
  \\
  Emma: 24-year-old ballet dancer striving for perfection and struggling with self-doubt.,
  \\
  \\
  Liam: 26-year-old musician with a carefree spirit and a charming andplayful nature.,

\end{tcolorbox}

\section{Examples for Scripts Crafting}
\label{sec:app:example_scripts_gen}
The following lists an example communication script from Human Communication Benchmark for "Andrew" and "Victoria" under "City Park Picnic" scene.
\begin{tcolorbox}[width=1\textwidth]

    $\{$
      "turn": 1,
      "role": Andrew,
      "style": default,
      "content": Hey Victoria, I'm glad we could finally enjoy this picnic together.
    $\}$,
    \\
  \\
    $\{$
      "turn": 2,
      "role": Victoria,
      "style": default,
      "content": Yes, Andrew. It's nice to take a break from work and spend time with friends.
    $\}$,\\
  \\
    $\{$
      "turn": 3,
      "role": Andrew,
      "style": happy,
      "content": Absolutely! So, how's your love life going lately? Any exciting stories to share?
    $\}$,\\
  \\
    $\{$
      "turn": 4,
      "role": Victoria,
      "style": sad,
      "content": Well, to be honest, it's been quite turbulent. I just can't seem to find the right person.
    $\}$,
    \\
  \\
    $\{$
      "turn": 5,
      "role": Andrew,
      "style": sympathetic,
      "content": I'm sorry to hear that. Don't worry, Victoria. The right person will come along. You just need to be patient.
    $\}$,
    \\
  \\
    $\{$
      "turn": 6,
      "role": Victoria,
      "style": default,
      "content": I know, but it's frustrating sometimes. I just want someone who understands my passion for journalism.
    $\}$,
    \\
  \\
    $\{$
      "turn": 7,
      "role": Andrew,
      "style": calm,
      "content": I can understand that. Being a journalist myself, it's important to have someone who can relate to your work.
    $\}$,
    \\
  \\
    $\{$
      "turn": 8,
      "role": Victoria,
      "style": angry,
      "content": Exactly! But most guys I meet either don't understand or feel threatened by my dedication.
    $\}$,
    \\
  \\
    $\{$
      "turn": 9,
      "role": Andrew,
      "style": sympathetic,
      "content": That's their loss, Victoria. You deserve someone who appreciates your drive and ambition.
    $\}$,
    \\
  \\
    $\{$
      "turn": 10,
      "role": Victoria,
      "style": sad,
      "content": I hope so, Andrew. Sometimes I feel like I'll never find that person.
    $\}$,
    \\
  \\
    $\{$
      "turn": 11,
      "role": Andrew,
      "style": happy,
      "content": Don't lose hope. Love has a mysterious way of finding us when we least expect it.
    $\}$,
    \\
  \\
    $\{$
      "turn": 12,
      "role": Victoria,
      "style": default,
      "content": You're right, Andrew. I'll keep searching and hope for the best.
    $\}$

\end{tcolorbox}

\clearpage
\section{Prompt for Consistency Score Evaluation}
\label{sec:app:prompt_consistency_score}
\begin{tcolorbox}[width=1\textwidth]
I will provide you with a scenario, characters, and scripts for the characters' communication within this scenario. Please evaluate and score the consistency between the script and the given scenario and characters. Specific requirements are as follows: When assessing the consistency between the script and the scenario, characters, delve into the key elements of each dialogue to ensure that the generated script closely aligns with the provided scenario, background, and character descriptions. First, focus on the descriptions of the scene and characters, evaluating whether the generated dialogue accurately presents the features of the scenario, including location, time, and atmosphere.\\
Below is the data:\\
$[$BEGIN DATA$]$\\
***\\
$[$scene$]$: $\{scene\}$\\
***\\
$[$roles$]$: $\{role\}$\\
***\\
$[$scripts$]$: $\{scripts\}$\\
***\\
$[$Criterion$]$: consistency:\\
"1": "Not consistent - The scripts is completely irrelevant with the provided scenario, characters, and dialogue."\\
"2": "Somewhat consistent - The scripts partially aligns with the provided scenario, characters, and dialogue. While some aspects are accurate, there are notable inconsistencies that affect the overall cohesion and believability of the response."\\
"3": "Moderately consistent - The scripts demonstrates a reasonable level of consistency with the provided scenario, characters, and dialogue. It generally aligns with the context, but there may be occasional lapses or minor discrepancies."\\
"4": "Consistent - The scripts is largely consistent with the provided scenario, characters, and dialogue. It effectively captures the essence of the context, providing a coherent and believable response. However, there might be a few minor inconsistencies that do not significantly impact the overall consistency."\\
"5": "Highly consistent - The scripts is exceptionally consistent with the provided scenario, characters, and dialogue. It accurately reflects the given context, maintaining a high level of coherence and believability throughout. There are no notable inconsistencies that detract from the overall consistency."\\
***\\
$[$END DATA$]$\\
Does the scripts meet the criterion? My score is: $[$insert score based on the provided consistency criteria$]$.
\end{tcolorbox}

\clearpage
\section{Prompt for Content Score Evaluation}
\label{sec:app:prompt_content_score}
\begin{tcolorbox}[width=1\textwidth]
I will provide you with a scripts for the multiple characters' communication. Please evaluate and score the quality and logical coherence of the script content. Specific requirements are as follows: Please conduct a thorough examination of each dialogue's language quality, emotional expression, logical consistency, and overall reasonableness. Begin by evaluating the language of each dialogue, ensuring that it is natural, fluent, and free from grammatical and lexical errors. Pay attention to emotional expression to ensure that the dialogues adequately convey the characters' emotions. Emphasize the assessment of logical coherence and reasonableness in the dialogues, ensuring that the characters' speech and actions align with common sense and that their decisions and behaviors possess sufficient rationale within the plot development.\\
Below is the data:\\
$[$BEGIN DATA$]$\\
***\\
$[$scripts$]$: $\{scripts\}$\\
***\\
$[$Criterion$]$: content quality and logical coherence:\\
"1": "Poor - The script lacks clarity, with language that is unclear or inappropriate. Emotional expression is poorly conveyed, and the dialogue lacks logical coherence, making it difficult to follow or believe."

"2": "Below average - The script demonstrates some clarity, but language usage may be inconsistent or contain errors. Emotional expression is present but may be inconsistent or not well conveyed. The logical coherence of the dialogue is compromised at times, affecting believability."

"3": "Average - The script generally maintains clarity in language, with few errors or inconsistencies. Emotional expression is reasonably conveyed, and there is a moderate level of logical coherence in the dialogue. However, some aspects may still lack depth or believability."

"4": "Above average - The script is clear and well-written, with minimal language issues. Emotional expression is effectively conveyed, and the dialogue exhibits a high level of logical coherence. The interactions and decisions of the characters are mostly believable, contributing to the overall quality."

"5": "Excellent - The script is exceptionally well-crafted with clear, engaging language. Emotional expression is vivid and effectively communicates the characters' feelings. The dialogue demonstrates outstanding logical coherence, ensuring that the characters' actions and decisions align seamlessly with the plot. The overall content quality is exceptional."

$[$END DATA$]$

Does the scripts meet the criterion? My score is: $[$insert score based on the provided content quality and logical coherence criteria$]$.
\end{tcolorbox}

\end{document}